\documentclass[10pt,twocolumn,letterpaper]{article}
\usepackage{wacv}
\usepackage{times}
\usepackage{epsfig}
\usepackage{graphicx}
\usepackage{amsmath}
\usepackage{amssymb}
\usepackage{booktabs}

\def\wacvPaperID{439} 

\wacvalgorithmstrack   

\wacvfinalcopy 

\usepackage{helvet} 
\usepackage{courier}  
\usepackage[utf8]{inputenc} 
\usepackage[T1]{fontenc}    
\usepackage{url}

\usepackage{booktabs}       
\usepackage{amsfonts}       
\usepackage{nicefrac}       
\usepackage{microtype}      

\usepackage{epsfig}
\usepackage{amsmath}
\usepackage{amssymb}
\usepackage{adjustbox}

\usepackage{amsthm}  
\usepackage{wasysym}
\usepackage[ruled,vlined,algo2e]{algorithm2e}

\usepackage{color}
\usepackage{dsfont}	
\usepackage{wrapfig}
\usepackage{footnote}
\usepackage{epstopdf}
 \usepackage{enumitem}
\usepackage{subfigure}
\usepackage{caption}
\usepackage{comment}
\usepackage{multirow}
\usepackage{algorithm}
\usepackage{algorithmic}
\usepackage{mathtools}
\usepackage{tcolorbox}

\def\eg{\emph{e.g., }}
\def\ie{\emph{i.e., }}

\def\etc{\emph{etc. }} 

\def\wrt{\emph{w.r.t. }}

\usepackage{pifont}

\usepackage{mathtools}

\makeatletter
\newcommand*{\rom}[1]{\expandafter\@slowromancap\romannumeral #1@}
\makeatother
\makeatletter
\newcommand\footnoteref[1]{\protected@xdef\@thefnmark{\ref{#1}}\@footnotemark}
\makeatother
\newcommand{\bfsection}[1]{\vspace*{0.1cm}\noindent\textbf{#1}}

\newcommand{\Poincare}{Poincar\'e\xspace}

\newcommand{\norm}[1]{\left\lVert#1\right\rVert}
\usepackage{amsmath}

 \newcommand{\squishlist}{
	\begin{list}{$\bullet$}
		{ \setlength{\itemsep}{0pt}
			\setlength{\parsep}{3pt}
			\setlength{\topsep}{3pt}
			\setlength{\partopsep}{0pt}
			\setlength{\leftmargin}{1.5em}
			\setlength{\labelwidth}{1em}
			\setlength{\labelsep}{0.5em} } }
	
	\newcommand{\squishlisttwo}{
		\begin{list}{$\bullet$}
			{ \setlength{\itemsep}{0pt}
				\setlength{\parsep}{0pt}
				\setlength{\topsep}{0pt}
				\setlength{\partopsep}{0pt}
				\setlength{\leftmargin}{2em}
				\setlength{\labelwidth}{1.5em}
				\setlength{\labelsep}{0.5em} } }
		
		\newcommand{\squishend}{
	\end{list}}

\ifwacvfinal
\usepackage[breaklinks=true,bookmarks=false]{hyperref}
\else
\usepackage[pagebackref=true,breaklinks=true,colorlinks,bookmarks=false]{hyperref}
\fi

\usepackage[accsupp]{axessibility}  

\begin{document}


\title{Understanding Hyperbolic Metric Learning through Hard Negative Sampling}

\author{Yun Yue$^{1}$
\and
Fangzhou Lin$^{1,2}$
\and 
Guanyi Mou$^{1}$ 
\and 
Ziming Zhang$^{1}$ \and 
$^1$Worcester Polytechnic Institute, USA \\
$^2$Tohoku University, Japan\\
{\tt\small \{yyue, flin2, gmou, zzhang15\}@wpi.edu}
}

\maketitle

\begin{abstract}
In recent years, there has been a growing trend of incorporating hyperbolic geometry methods into computer vision. While these methods have achieved state-of-the-art performance on various metric learning tasks using hyperbolic distance measurements, the underlying theoretical analysis supporting this superior performance remains under-exploited. In this study, we investigate the effects of integrating hyperbolic space into metric learning, particularly when training with contrastive loss. We identify a need for a comprehensive comparison between Euclidean and hyperbolic spaces regarding the temperature effect in the contrastive loss within the existing literature. To address this gap, we conduct an extensive investigation to benchmark the results of Vision Transformers (ViTs) using a hybrid objective function that combines loss from Euclidean and hyperbolic spaces. Additionally, we provide a theoretical analysis of the observed performance improvement. We also reveal that hyperbolic metric learning is highly related to hard negative sampling, providing insights for future work. This work will provide valuable data points and experience in understanding hyperbolic image embeddings. To shed more light on problem-solving and encourage further investigation into our approach, our code \footnote{\url{https://github.com/YunYunY/HypMix}} is available online.

\end{abstract}

\section{Introduction}
In computer vision, the central concept of metric learning~\cite{musgrave2020metric} is to bring similar data representations (positive pairs) closer in the embedding space while separating dissimilar data representations (negative pairs). This approach hinges on the embedding space's similarity, which mirrors semantic similarity. Applications of metric learning include content-based image retrieval~\cite{hoi2010semi,li2015weakly,NPair,ProxyNCA,PNCAPP,sop}, near-duplicate detection~\cite{zheng2016improving,kordopatis2017near,he2019part,jiang2019svd}, face recognition~\cite{cao2013similarity,huang2015projection,sphereface,facenet}, person re-identification~\cite{yi2014deep,liao2015efficient,chen2017triplet,xiao2017joint}, zero-shot~\cite{sop,bucher2016improving,jiang2019adaptive,chen2019hybrid} and few-shot learning~\cite{snell2017prototypical,sung2018learning,qiao2019transductive,jiang2020multi}.

\begin{figure}
\begin{center}
   \includegraphics[width=1.\linewidth]{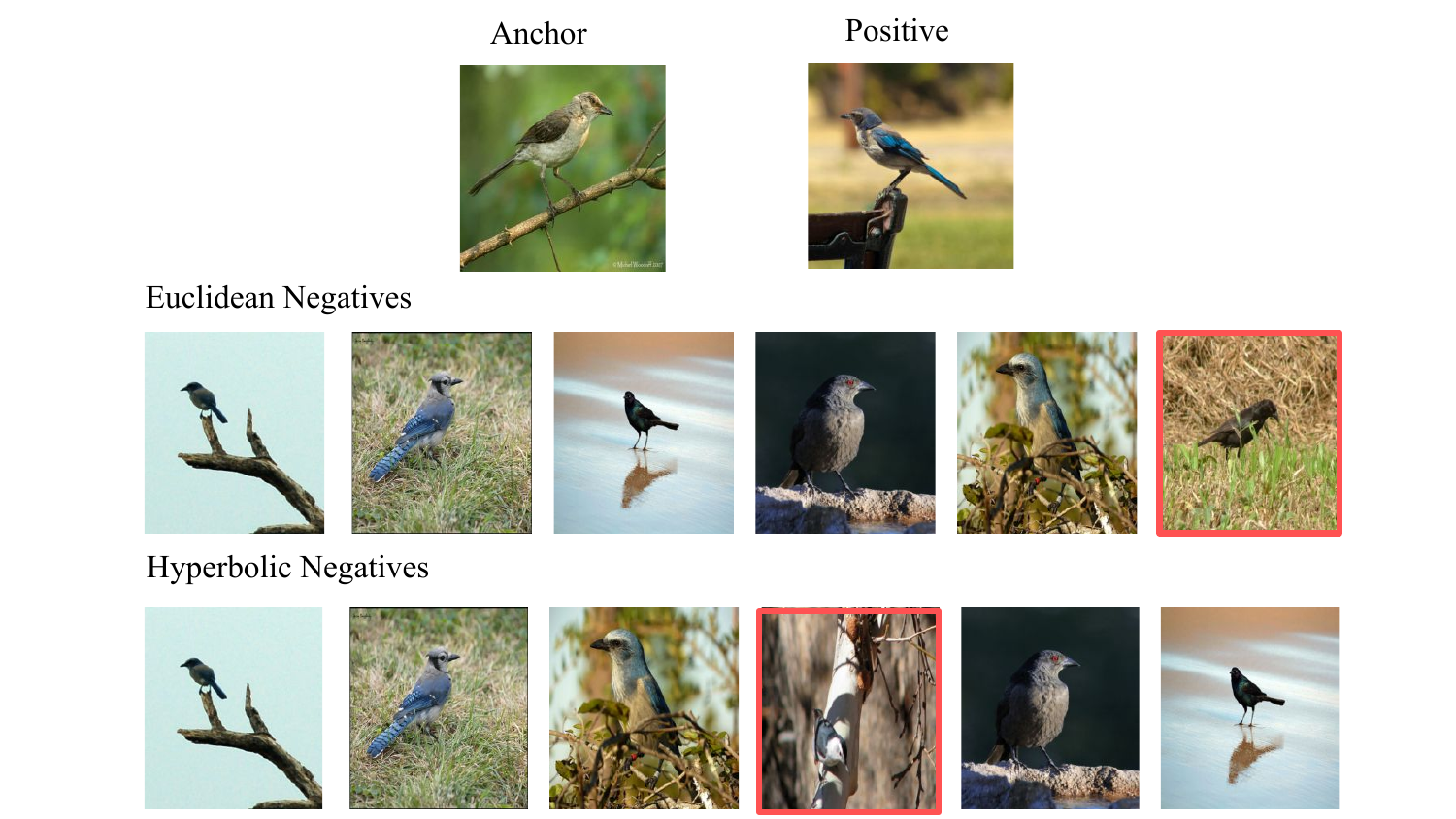}
\end{center}
    \vspace{-3.0mm}

    \caption {For a random chosen anchor, we arrange hard negatives selected by a well-trained model based on their distance to the anchor. The top row displays the anchor and its positive pair. We present sorted negatives from Euclidean embedding model (``Sph-ViT") and the hyperbolic embedding model (``Hyp-ViT"), respectively. Negatives are ordered by increasing distance from left to right. ``Sph-ViT" is trained with $\tau=0.05$, while ``Hyp-ViT" is trained with $\tau=0.05$ and $c=0.1$. Red boxes highlight negatives that are absent in the top 6 hard negatives of the other model.}
    \vspace{-3.0mm}

\label{fig:negative}
\end{figure}

Pair-wise losses including contrastive loss \cite{hadsell2006dimensionality}, triplet loss \cite{weinberger2009distance}, N-pair
loss \cite{sohn2016improved}, and InfoNCE \cite{oord2018representation, chen2020simple} are fundamental in metric learning, constructed based on instance relationships within a batch.
Contrastive loss, for instance, treats views of data from the same class as positive pairs and other batch data as negative pairs, encouraging the encoder to map positive pairs close and negative pairs apart. Contrastive learning embeddings are positioned on a hypersphere using the inner product as a distance measure \cite{chen2021intriguing}. However, fields like social networks, brain imaging, and computer graphics often exhibit hierarchical structures \cite{bronstein2017geometric}.

A recent trend in representation learning is using hyperbolic spaces, which excel in capturing hierarchical data \cite{peng2021hyperbolic}. Unlike Euclidean space with polynomial volume growth, hyperbolic space demonstrates exponential growth suitable for tree-like data structures. Hyperbolic representations have shown success in NLP \cite{nickel2017poincare, nickel2018learning}, image segmentation \cite{weng2021unsupervised, atigh2022hyperbolic}, few-shot \cite{khrulkov2020hyperbolic} and zero-shot learning \cite{liu2020hyperbolic}, as well as representation learning and metric learning \cite{ermolov2022hyperbolic, ge2022hyperbolic, yue2023hyperbolic, liu2022enhancing, lin2022contrastive}. In hyperbolic space, feature similarity is calculated via distance measurement after projecting data features.

\noindent
\textbf{Motivation.} While hyperbolic contrastive loss is effective in metric learning, the underlying reason has yet to be fully understood. Prior research has demonstrated that employing a vision transformer-based model for metric learning with output embeddings mapped to hyperbolic space outperforms Euclidean embeddings \cite{ermolov2022hyperbolic}. Nonetheless, our experimental findings indicate that a straightforward embedding approach in one geometry doesn't consistently yield superior results compared to the other. The choice of temperature parameter ($\tau$) in the loss function and the curvature values in hyperbolic space contribute to the performance variations among different embeddings. Moreover, Euclidean and hyperbolic embeddings exhibit complementary behavior with diverse $\tau$ and curvature values. As depicted in Fig.~\ref{fig:negative}, embedding features under different geometries occasionally highlight different crucial instances. This observation sparks our interest in delving further into hyperbolic metric learning.

 The pair-wise contrastive loss relies on two key components: positive pairs $(\mathbf x, \mathbf x^+)$, and negative pairs $(\mathbf x, \mathbf x^-)$ of data. The strategies of positive sampling  has been studied intensively \cite{blum1998combining, xu2013survey, bachman2019learning, chen2020simple, tian2020makes, chen2020improved, tian2020contrastive, srinivas2020curl, logeswaran2018efficient, oord2018representation, purushwalkam2020demystifying, sermanet2018time}, as well as the negative data augmentation \eg \cite{kalantidis2020hard, ge2021robust,  robinson2021contrastive, sinha2021negative} where most of the works focus on ``hard'' negative data augmentation. For instance, \cite{robinson2021contrastive} provided a popular principle that ``{\em The most useful negative samples are ones that the embedding currently believes to be similar to the anchor}'', where the embedding refers to the network output. That is, letting $\mathbf x_1^-, \mathbf x_2^-$ be two negative samples \wrt the anchor $\mathbf x$ and $\phi$ be the current network, then $x_1^-$ is more useful (\ie harder) than $\mathbf x_2^-$ if $\phi(\mathbf x)^T\phi(\mathbf x_1^-) > \phi(\mathbf x)^T\phi(\mathbf x_2^-)$ holds, where $(\cdot)^T$ denotes the matrix transpose operator.

 Our analysis reveals that the differences in embedding performance stem from the distinct effects of hard negative sampling in the two geometries. This analysis shifts our focus from individual hard negative instances to hard triplets. We demonstrate that triplet selection hinges on a weight $p(\mathbf x^-)$, which varies across different geometries for the same instance. To capitalize on the benefits of these diverse geometries, we introduce the concept of embedding fusion. Our solution is driven by two essential insights: 1) different geometries yield variety in hard negative selection properties, and 2) distinct geometries complement each other in triplet selection.

In contrast to prior studies that emphasize selecting more meaningful negative samples \cite{chuang2020debiased, robinson2020contrastive, wang2021understanding, kalantidis2020hard, zhang2017mixup, hu2021adco, shah2022max, ge2021robust}, our primary contribution lies in understanding the performance disparity across different geometries. Our experiments unveil the complementary performance of diverse geometries. We delve into unexplored aspects of hyperbolic metric learning through an analysis of hard triplet selection. Furthermore, we present a direct solution for effectively utilizing information from different geometries. Demonstrating the efficacy of our straightforward fusion algorithm, we adhere to the standard design of vision transformer-based metric learning. By adopting an ensemble approach, our model effectively captures more informative negative samples from an additional pool, resulting in enhanced performance.

\section{Related Work}
\bfsection{Hyperbolic Embeddings.}
The machine learning community has a long history of embracing Euclidean space for representation learning. It is a natural generalization of our intuition-friendly, physically-accessible three-dimensional space where the measurable distance is represented with inner-product~\cite{ganea2018hyperbolic, peng2021hyperbolic}. 
However, the Euclidean embedding may not best fit in some complex tree-like data fields such as Biology, Network Science, Computer Graphics, or Computer Vision that exhibit highly non-Euclidean latent geometry \cite{ganea2018hyperbolic, bronstein2017geometric}. 

In the computer vision domain, the hyperbolic space has been found well-suited for image segmentation~\cite{weng2021unsupervised, atigh2022hyperbolic}, zero-shot recognition~\cite{liu2020hyperbolic, fang2021kernel}, few-shot image classification~\cite{khrulkov2020hyperbolic, fang2021kernel, gao2021curvature} as well as point cloud classification~\cite{montanaro2022rethinking}. The work of \cite{guo2022clipped} revealed the vanishing gradients issue of Hyperbolic Neural Networks (HNNs) when applied to classification benchmarks that may not exhibit hierarchies and showed clipped HNNs are more robust to adversarial attacks. 
Concurrently, \cite{ermolov2022hyperbolic, yue2023hyperbolic, ge2022hyperbolic} mapped the output-of-image representations encoded by a backbone to a hyperbolic space so that the representations of similar objects in the embedding space are pulled together. They empirically verified the performance of pairwise cross-entropy loss with the hyperbolic distance in image embeddings while we investigate the reason why contrastive metric learning works in hyperbolic space and when the method will work without hyperbolic embedding.  

\bfsection{Pair-wise Losses.}
One of the fundamental losses in metric learning is contrastive loss \cite{hadsell2006dimensionality}. The contrastive representation learning attempts to pull the embeddings of positive pairs closer and push the embeddings of negative pairs away in the latent space by optimizing the pair-wise objective. Following these fundamental concepts, many variants have since been proposed, such as triplet margin loss \cite{weinberger2009distance}, angular loss \cite{wang2017deep}, margin loss \cite{wu2017sampling}, \etc Other losses contain softmax operation and LogSumExp for a smooth approximation of the maximum function. For example, the lifted structure loss \cite{oh2016deep} applies LogSumExp to all negative pairs, and the N-Pairs loss \cite{sohn2016improved} applies the softmax function to each positive pair relative to all other pairs. Recently, learning representations from unlabeled data in contrastive way \cite{chopra2005learning, hadsell2006dimensionality} has been one of the most competitive research field \cite{oord2018representation, hjelm2018learning, wu2018unsupervised, tian2020contrastive, sohn2016improved, chen2020simple, jaiswal2020survey, li2020prototypical, he2020momentum, chen2020improved, chen2020big, bachman2019learning, misra2020self, caron2020unsupervised}. Popular model structures like SimCLR \cite{chen2020simple} and Moco \cite{he2020momentum} apply the commonly used loss function InfoNCE \cite{oord2018representation} to learn a latent representation that is beneficial to downstream tasks. Several theoretical studies show that self-supervised contrastive loss optimizes data representations by aligning positive pairs while pushing negative pairs away on the hypersphere \cite{wang2020understanding, chen2021intriguing, wang2021understanding, arora2019theoretical}. Recently, the loss has been applied in metric learning and shows superior performance when equipped with ViT and hyperbolic embedding. 

\bfsection{Hard-negative Mining.}
When applying contrastive loss, the positive pairs could be different modalities of a signal \cite{arandjelovic2018objects, tian2020contrastive, tschannen2020self}, different data augmentations of the same image, \eg{color distortion, random crop} \cite{chen2020simple, chen2020improved,grill2020bootstrap} or instances from the same category like in image retrieval. \cite{tian2020makes} suggested generating the positive pairs with “InfoMin principle" so that the generated positive pairs maintain the minimal information necessary for the downstream tasks. \cite{selvaraju2021casting, peng2022crafting, mishra2021object, li2022univip} proposed selecting meaningful but not fully overlapped contrastive crops with guidance like attention maps or object-scene relations. \cite{shen2020mix} empirically demonstrated that introducing extra convex combinations of data as positive augmentation improves representation learning. Similar mixing data strategies could be found in \cite{lee2020mix, kim2020mixco, verma2021towards,li2020self,ren2022sdmp}. Other than exploring positive augmentation, recent works focus on negative data selection in contrastive learning. Typically, negative samples are drawn uniformly from the training data. Based on the argument that not all negatives are true negatives, \cite{chuang2020debiased, robinson2020contrastive} developed debiased contrastive loss to assign higher weights to the hard negative samples. \cite{wang2021understanding} proposed an explicit way to select the hard negative samples similar to the positive. To provide more meaningful negative samples, \cite{kalantidis2020hard} studied the Mixup \cite{zhang2017mixup} strategy in latent space to generate hard negatives. \cite{hu2021adco} proposed directly learning a set of negative adversaries. \cite{ge2021robust} generated negative samples by texture synthesis or selecting non-semantic patches from existing images. Unlike previous studies, we do not propose a new method for negative data sampling but provide some insights on the real ``hard'' negatives from the perspective of the gradients of contrastive loss. We try to understand when data are embedded in different spaces, which factor contributes to the ``good'' embeddings. Based on our analysis, we propose to learn embeddings under a mixture of geometries.  

\section{Understanding Geometry Effect}

We introduce the relevant preliminaries, such as hyperbolic geometry and pairwise cross-entropy used in metric learning, in Section \ref{sec:method-hyper}. The relationship between geometries and triplet selection is analyzed in Section \ref{sec:method-analysis}. In Section \ref{sec:method-mix}, we present embedding fusion as a strategy to utilize complementary information from diverse geometries, accompanied by discussing the distinctions between our approach and existing hard negative sampling methods.

\subsection{Preliminaries}\label{sec:method-hyper}

Among several isometric models \cite{cannon1997hyperbolic} of hyperbolic space,
we stick to the \Poincare ball model \cite{sarkar2012low} that is well-suited for gradient-based optimization (\ie the distance function is differentiable). In particular, the model $(\mathbb{M}^n_c, g^{\mathbb{M}})$ is defined by the manifold $\mathbb{M}^n = \{ x \in \mathbb{R}^n \colon c\|x\|^2 < 1, c \geq 0\} $ equipped with the Riemannian metric $g^{\mathbb{M}} = \lambda_c^2 g^E$, where $c$ is the curvature parameter and $ \lambda_c = \frac{2}{1-c\|x\|^2}$ is conformal factor that scales the local distances. $g^E = \mathbf{I}_n$ denotes the Euclidean metric tensor. 

The framework of \emph{gyrovector spaces} provides an elegant non-associative algebraic formalism for hyperbolic geometry just as vector spaces provide the algebraic setting for Euclidean geometry \cite{cannon1997hyperbolic, ungar2008analytic, ungar2008gyrovector, ganea2018hyperbolic}. For two vectors $\mathbf{x}, \mathbf{y} \in \mathbb{M}^n_c$, their addition is defined as
\begin{equation}\label{eq:add}
\small
    \mathbf{x} \oplus_c \mathbf{y} = \frac{(1+2c\langle \mathbf{x}, \mathbf{y} \rangle + c\|\mathbf{y}\|^2) \mathbf{x}+ (1-c\|\mathbf{x}\|^2)\mathbf{y}}{1+2c\langle \mathbf{x}, \mathbf{y} \rangle + c^2 \|\mathbf{x}\|^2 \|\mathbf{y}\|^2}.
\end{equation}
The hyperbolic distance between $\mathbf{x}, \mathbf{y} \in \mathbb{M}^n_c$ is defined as:
\begin{equation}\label{eq:hdist}
\small
    D_{hyp}(\mathbf{x},\mathbf{y}) = \frac{2}{\sqrt{c}} \mathrm{arctanh}(\sqrt{c}\|-\mathbf{x} \oplus_c \mathbf{y}\|).
\end{equation}
In particular, when $c = 0$, the Eq. \ref{eq:add} is the Euclidean addition of two vectors in $\mathbb{R}^n$ and Eq. \ref{eq:hdist} recovers Euclidean geometry: $\lim_{c \to 0} D_{hyp}(\mathbf{x},\mathbf{y})=2\|\mathbf{x}-\mathbf{y}\|.$ For an open $n$-dimensional unit ball, the geodesics of the \Poincare disk are then circles that are orthogonal to the boundary of the ball.

Before performing operations in the hyperbolic space, a map termed \emph{exponential} is used when mapping from the Euclidean space to the hyperbolic geometry.\cite{khrulkov2020hyperbolic}. 
The \emph{exponential} map is defined as:
\begin{equation}\label{eq:exp}\small
    \exp_\mathbf{x}^c(\mathbf{v}) = \mathbf{x} \oplus_ c \bigg(\tanh \bigg(\sqrt{c} \frac{\lambda_\mathbf{x}^c \|\mathbf{v}\|}{2} \bigg) \frac{\mathbf{v}}{\sqrt{c}\|\mathbf{v}\|}\bigg)
\end{equation}
In practice, we follow the setting of \cite{khrulkov2020hyperbolic} and \cite{ermolov2022hyperbolic} with the base point $\mathbf{x}=\mathbf{0}$ so that the formulas are less cumbersome and empirically have little impact on the obtained results.

For metric learning method, we adapt the approach suggested by \cite{ermolov2022hyperbolic}, each iteration involves sampling two data points, $(\mathbf x, \mathbf x^+)$, from $N$ distinct image categories to form positive pairs. All the other data in the same batch construct negative pairs with the anchor $(\mathbf x, \mathbf x^-)$. In this case, the total number of samples (batch size) is $K = 2N$ consisting of $N$ positive pairs. 

\bfsection{Pairwise Cross-Entropy Loss.}
Self-supervised contrastive learning, such as \cite{chen2020simple,tian2019contrastive,hjelm2018learning}, employs the following InfoNCE loss to attract positive pairs and separate negatives from the anchor in the latent space.

\begin{equation}
\small
  \mathcal{L}_{NCE}
  =-\sum_{i\in I}\log{
  \frac{\text{exp}\left(s_{i, i}/ \tau\right)}{\text{exp}\left(s_{i, i}/ \tau\right) + \sum_{k \neq i}\text{exp}\left( s_{i, k}/\tau\right)}
  }
  \label{eqn:self_loss}
\end{equation} 
In this context, $(i,i)$ refers to the anchor and its positive pair, and the other $2(N-1)$ indices denote the anchor's negatives. For each anchor $i$, there exists $1$ positive pair and $2N - 2$ negative pairs. The denominator comprises a total of $2N - 1$ terms. $\tau\in\mathcal{R}^+$ is the scalar temperature parameter that governs sharpness, $s_{i, j} = \mathbf{z}_i\cdot\mathbf{z}_j$ with $\mathbf{z}=g(f(\mathbf{x}))$. $f(\cdot)$ maps $\mathbf{x}$ to lower dimension using a shared ViT architecture \cite{dosovitskiy2020image}. $g(\cdot)$ projects $f(\mathbf{x})$ to latent space. Typically, $\mathbf{z}$ is normalized before loss calculation to lie on a unit hypersphere. 
\begin{equation}
\small
    D_{cos}(\mathbf{z}_i, \mathbf{z}_j)  = \norm{ \frac{\mathbf{z}_i}{\norm{\mathbf{z}_i}_2} - \frac{\mathbf{z}_j}{\norm{\mathbf{z}_j}_2} }^2_2 
 = 2 - 2 \frac{ \mathbf{z}_i \cdot \mathbf{z}_j }{\norm{\mathbf{z}_i}_2 \cdot \norm{\mathbf{z}_j}_2}
\label{eq:dcos}
\end{equation}
%
The general loss form of pairwise cross-entropy loss in Euclidean and hyperbolic space is defined as    
\begin{equation}
\small
\resizebox{0.85\hsize}{!}{%
  $\mathcal{L}
  =-\sum_{i\in I}\log{
  \frac{\text{exp}\left(-D_{i, i}/\tau\right)}{\text{exp}\left(-D_{i, i}/\tau\right) + \sum_{k \neq i}\text{exp}\left(-D_{i, k}/\tau\right)}
  }$
}
  \label{eqn:hcl}
\end{equation} 
where $D_{i,j}$ is the distance measurement like $D_{cos}(\mathbf{z}_i, \mathbf{z}_j)$ or $D_{hyp}(\mathbf{z}_i, \mathbf{z}_j)$. 

\subsection{Geometries vs. Hard Negatives}\label{sec:method-analysis}
\begin{figure*}
\begin{center}
   \includegraphics[width=0.8\linewidth]{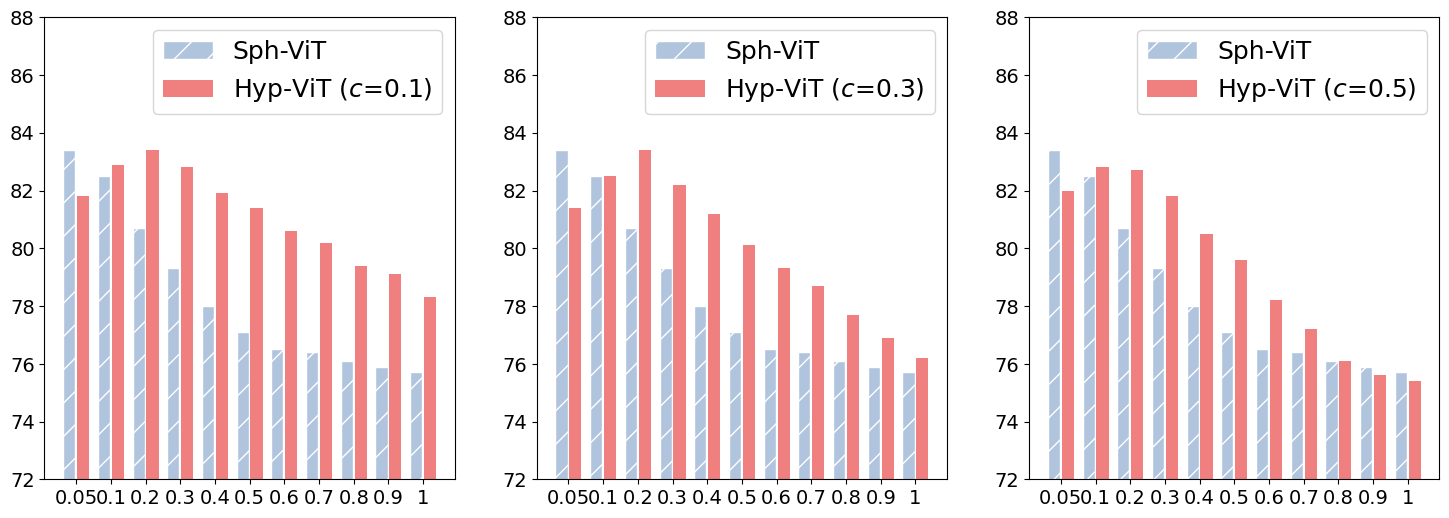}
\end{center}
 \caption {Experimentally, we find that Euclidean (``Sph'') and hyperbolic (``Hyp'') embeddings, with varying $\tau$ and $c$ values, exhibit complementary characteristics which is attributed to differences in negative selection across distinct geometry embeddings. To leverage this complementary information from different geometries, we introduce embedding fusion.} 
\label{fig:bar}
\end{figure*}

In this study, we investigate the conditions under which contrastive loss yields superior performance across distinct geometries. Fig. \ref{fig:bar} presents the performance comparison of ViT models trained with varying temperature values ($\tau$) and curvature parameter ($c$). \mbox{``Sph-''} are versions with Euclidean ($c=0$) embeddings optimized using $D_{cos}$ (\ref{eq:dcos}). ``Hyp-'' are versions with hyperbolic embeddings optimized using $D_{hyp}$ (\ref{eq:hdist}). Hyperbolic embeddings generally enhances performance, especially when $\tau \geq 0.2$ as shown in Fig. \ref{fig:bar}. From a hyperbolic geometry perspective, in hyperbolic space, the distance between a point and the origin experiences exponential growth. When comparing two points, the relative distance between them becomes larger in hyperbolic space due to this exponential distance growth. In the following analysis, we will demonstrate how this geometry impacts negative selection.

Our analysis is supported by empirical experiments utilizing the current state-of-the-art approach in metric learning for image retrieval.
Through gradient analysis of the InfoNCE loss in Euclidean space, followed by its generalization to hyperbolic geometry, we reveal the substantial impact of various geometries on the triplet selection weights $(\mathbf x, \mathbf x^+, \mathbf x^-)$.

Specifically, $\mathcal{L}_{NCE}$ could be further rewritten as 
\begin{equation}\label{eqn: loss2}
\small
\begin{split}
    \mathcal{L}_{NCE} & =  \sum_{i\in I}\log  \left( {1 + \frac{ \sum_{k \neq i}\text{exp}\left( s_{i, k}/\tau\right) }{\text{exp}\left(s_{i, i}/ \tau\right)}} \right) \\
    & =  \sum_{i\in I}\log  \left(  1 + \sum_{k \neq i} \text{exp} \left( \frac{1}{\tau}\left( s_{i, k} - s_{i, i}\right) \right)\right).
\end{split}
\end{equation}
The gradient is derived as: $\nabla\mathcal{L}_{NCE} = \sum_{i\in I} \nabla\mathcal{L}_{i} $ where

\begin{equation}\label{eq:grad}\small
    \nabla\mathcal{L}_{NCE_i} = \frac{1}{\tau} \sum_{k \neq i} p(\mathbf x^-) \nabla \left ( s_{i,k} - s_{i, i}\right)
\end{equation}
with $p(\mathbf x^-) =   \frac{\text{exp} \left(  s_{i, k} / \tau \right )}{\text{exp} \left(  s_{i, i} / \tau \right ) + \sum_{k \neq i} \text{exp} \left(  s_{i, k} / \tau \right )} \in[0,1] $. Detailed derivations are presented in the supplementary material.

Then stochastic gradient descent (SGD) can be used to update network weights. 
As for the general gradient form that includes hyperbolic space, it is just 
\begin{equation} \label{eq:grad_hyp}\small
    \nabla\mathcal{L}_{hyp_i} =\frac{1}{\tau} \sum_{k \neq i} p(\mathbf x^-) \nabla \left (D_{i, i} - D_{i,k}\right)
\end{equation}
with  $p(\mathbf x^-) =   \frac{\text{exp} \left(  - D_{i, k} / \tau \right )}{\text{exp} \left(  -D_{i, i} / \tau \right ) + \sum_{k \neq i} \text{exp} \left(  - D_{i, k} / \tau \right )} \in[0,1] $.\\

The gradient analysis shows that each $p(\mathbf x^-)$ decides the weights contributed by each triplet $(\mathbf x, \mathbf x^+, \mathbf x^-)$ to the gradient update. $p(\mathbf x^-)$ is further related to the relative distance between positve pair and negative pairs. By looking into the term $p(\mathbf x^-)$ we get the following conclusions: 
\squishlist
 \item $p(\mathbf x^-)$ is non-uniform for all triplet selection. Temperature parameter $\tau$ plays an important role in deciding the weights of gradient update for each negative pair. $p(\mathbf x^-)$ is larger with smaller $\tau$.
 \item The relative distance between $(\mathbf x, \mathbf x^+)$ and $(\mathbf x, \mathbf x^-)$ decides the gradient contribution by a negative instance.  
 \item For the same $\tau$, when we change from Euclidean distance to hyperbolic distance (varying $c$), the relative distance between a negative pair and a positive pair will be enlarged due to the geometry property of hyperbolic space. This will further affect the hard negative sampling.
\squishend

\begin{figure}[th]
    \begin{center}
	\centerline{\includegraphics[width=1\linewidth, keepaspectratio,]
        {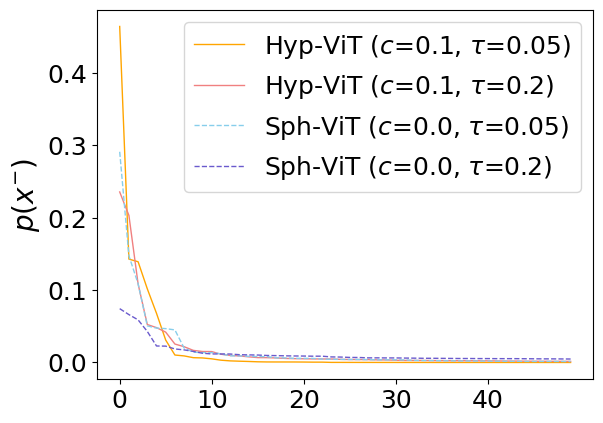}}
	\end{center}
	\vspace{-20pt}
    \caption{$p(\mathbf x^-)$ for different ViT. x-axis is the index of data points when the distance between $(\mathbf x^-)$ to anchor point $\mathbf x$ is sorted in ascending order (better view in color print).} 
\label{fig:p}
	\vspace{-3mm}
\end{figure}

\subsection{Ensemble Learner with Mix Geometries}\label{sec:method-mix}

We empirically verify the above analysis with the method proposed by \cite{ermolov2022hyperbolic}. We follow the exact experiment settings proposed by previous work. The detail of the experiments can be found in the following section.

As asserted by \cite{ermolov2022hyperbolic}, optimizing pairwise contrastive loss in hyperbolic space yields a new state-of-the-art performance surpassing Euclidean embeddings. In their study, they set $\tau = 0.1$ for Euclidean space and $\tau = 0.2$ for hyperbolic space, with a curvature parameter of $c=0.1$. Interestingly, upon conducting similar experiments with varying $\tau$ from a small value of $0.05$ to $1$, we observe that proper tuning of $\tau$ can lead to optimal performance in Euclidean space embeddings as well. Both Euclidean and hyperbolic embeddings exhibit performance drops with larger $\tau$ values. This trend is consistent across different backbone encoders, as illustrated in the supplemental material. The impact of $\tau$ relates to $p(\mathbf x^-)$ in Eq. \ref{eq:grad} and Eq. \ref{eq:grad_hyp}, where smaller $\tau$ assigns higher weights to harder negatives (i.e., negative samples $\mathbf x^-$ that are closer to the anchor $\mathbf x$ than the positive sample $\mathbf{x}^+$).

We empirically validate our analysis of $p(\mathbf x^-)$ by visualizing a small batch $\mathbf x ^-$ after the model converges (Fig. \ref{fig:p}). The horizontal axis represents data point indices, sorted based on their distance from the anchor point. We present the results for two sets of temperature values: $\tau=0.05$ for Euclidean embeddings and $\tau=0.2$ for hyperbolic embeddings, corresponding to their respective optimal performance points. Notably, at $\tau=0.05$, the weights $p(\mathbf x^-)$ for hard negatives can reach as high as 0.46, compared to Euclidean embeddings. Similarly, at $\tau=0.2$, instances closer to the anchor exhibit significantly higher $p(\mathbf x^-)$ than in Euclidean embeddings. This observation confirms the non-uniform distribution of $p(\mathbf x^-)$. Among all available triplet combinations ($\mathbf x, \mathbf x ^+, \mathbf x ^-$), the majority exhibit small $p(\mathbf x^-)$ values, while a small portion has relatively larger values. It is these selected triplets that significantly contribute to shaping the decision boundary. The process of selecting important triplets is greatly influenced by $p(\mathbf x^-)$, which plays a crucial role. Different geometries lead to distinct feature selections, guided by the variations in $p(\mathbf x^-)$.

Motivated by our observation of complementary performance across different geometries, the idea of fusing their features to harness this complementary information arises naturally. In Fig. \ref{fig:negative}, we illustrate the hard negatives selected by well-trained models embedded in different geometries, based on their distances to the anchor. The distinct triplet pools selected by these embeddings are evident. For the same instance, $p(\mathbf x^-)$ varies between Euclidean and hyperbolic embeddings. If both $p(\mathbf x^-)$ values are small, the instance doesn't significantly contribute to the decision boundary. If one value is high, it will be considered if selected from the fusion pool.

\bfsection{Learning with Mixed Geometries.} As our analysis indicates, within each embedding space, the model effectively selects hard negative samples for proficient representation learning. Fusion empowers the model to choose from the amalgamated triplet pool. Two direct approaches exist for constructing the ensemble learner. One involves a convex combination of loss functions from two distinct geometries, such as $ \lambda \mathcal{L}_{hyp} + (1 - \lambda )\mathcal{L}_{NCE}$. Yet, our experimental tests indicated that this form did not yield improved results. Consequently, we adopt the second approach, which involves feature space fusion.
Our proposed $ \mathcal{L}_{mix}$ is defined as 
\begin{equation}
\small
\resizebox{0.9\hsize}{!}{%
$\mathcal{L}_{mix} = -\sum_{i\in I}\log{
  \frac{\text{exp}\left(\left (-D_{cos}(\mathbf{z}_i, \mathbf{z}_{i}) - \lambda D_{hyp}(\mathbf{z}_i, \mathbf{z}_{i}) \right )/\tau\right)}{\sum \text{exp}\left ( \left(-D_{cos}(\mathbf{z}_i, \mathbf{z}_k) - \lambda D_{hyp}(\mathbf{z}_i, \mathbf{z}_a) \right )/\tau\right)}
  }$
  }
  \label{eqn:mix}
\end{equation} 
where $\lambda$ is a tunable hyperparameter that controls the hard negative selecting effect from different geometries.

\bfsection{Comparing with Hard Negative Sampling Methods.}
Recent studies have concentrated on negative data selection within contrastive learning. Notably, \cite{chuang2020debiased, robinson2020contrastive} introduced debiased contrastive loss to assign higher weights to challenging negative samples. Alternatively, \cite{wang2021understanding} proposed an explicit strategy for selecting hard negative samples that resemble positives. To enhance the relevance of negative samples, \cite{kalantidis2020hard} explored Mixup \cite{zhang2017mixup} in the latent space to generate challenging negatives. \cite{hu2021adco} directly learned negative adversaries, while \cite{shah2022max} proposed negative selection through quadratic optimization, and \cite{ge2021robust} generated negatives via texture synthesis or non-semantic patches.

Diverging from previous works, our study comprehends hyperbolic geometry through the lens of triplet selection. We do not solely aim to enhance contrastive learning via hard negative sampling. Rather, we scrutinize the assertion that hyperbolic embeddings consistently outperform Euclidean embeddings, as shown in earlier studies. Our analysis reveals that triplet selection varies across geometries. From this perspective, we propose that combining different geometries can expand the sampling pool. Our experiments demonstrate that mixed geometries prompt the model to select triplets crucial to the decision boundary, while discarding less significant ones present in both embeddings. A potential future direction could be merging existing negative sampling methods with different geometries.


\setlength{\tabcolsep}{0.47em}
\begin{table*}[htbp]
  \centering
  \small
  \scalebox{.85}{
  \begin{tabular}{l|cccc|cccc|cccc}
    \toprule
    \multirow{2}{*}{Method} &
    \multicolumn{4}{c|}{CUB-200-2011 (K)} &
    \multicolumn{4}{c|}{Cars-196 (K)} &
    \multicolumn{4}{c}{SOP (K)} \\
    &
    1 & 2 & 4 & 8 &
    1 & 2 & 4 & 8 &
    1 & 10 & 100 & 1000 \\
    \midrule
    Margin \cite{Margin} &
    63.9 & 75.3 & 84.4 & 90.6 &
    79.6 & 86.5 & 91.9 & 95.1 &
    72.7 & 86.2 & 93.8 & 98.0\\
    FastAP \cite{FastAP} &
    - & - & - & - &
    - & - & - & - &
    73.8 & 88.0 & 94.9 & 98.3\\
    NSoftmax \cite{NSoftmax} &
    56.5 & 69.6 & 79.9 & 87.6 &
    81.6 & 88.7 & 93.4 & 96.3 &
    75.2 & 88.7 & 95.2 & - \\
    MIC \cite{MIC} &
    66.1 & 76.8 & 85.6 & - &
    82.6 & 89.1 & 93.2 & - &
    77.2 & 89.4 & 94.6 & - \\
    XBM \cite{XBM} &
    - & - & - & - &
    - & - & - & - &
    80.6 & 91.6 & 96.2 & 98.7 \\
    IRT\textsubscript{R} \cite{IRT} &
    72.6 & 81.9 & 88.7 & 92.8 &
    - & - & - & - &
    83.4 & 93.0 & 97.0 & 99.0\\
    \midrule
    Sph-DINO &
    77.9 & 86.1 & 91.6 & 95.1 &
    82.5 & 89.2 & 93.3 & 95.8 & 
    84.6 & 94.2 & 97.6 & 99.2 
    \\
    Sph-ViT \textsuperscript{$\mathsection$} &
    83.4 & 90.2 & 94.0 &  96.3 & 
    78.8 & 86.7 & 91.9 & 95.0 &
    85.5 & {\bf 94.9} & {\bf 98.2} & {\bf 99.5} 
    \\
    Hyp-DINO &
    78.3 & 86.8 & 92.0 & 95.4 & 
    83.2 & 90.0 & 93.8 & 96.3 &
    84.0 & 93.9 & 97.6 & 99.2 
    \\
    Hyp-ViT \textsuperscript{$\mathsection$} &
    83.4 & 90.2 & {\bf 94.2} & 96.3 &
    80.1 & 88.1 & 92.8 & 95.7 & 
    85.3 & 94.8 & 98.1 & {\bf 99.5} 
   \\
    Mix-DINO &
    78.5 & 86.9 & 91.8 & 95.3 & 
    {\bf 85.2} & {\bf 91.2} &{\bf 94.6} & {\bf 96.8} & 
    84.9 & 94.3 & 97.7 & 99.3 
   \\
    Mix-ViT \textsuperscript{$\mathsection$} &
    {\bf 83.7} & {\bf 90.4} & {\bf 94.1} & {\bf 96.5} &
    81.3 &  88.5 & 93.2 &  96.1 & 
    {\bf 85.8} & {\bf 95.0} & {\bf 98.1} & {\bf 99.5}
   \\
    \bottomrule

  \end{tabular}}
    \caption{Recall@K for different datasets with embeddings $dim=128$. Our methods are in the bottom two rows, evaluated for head embeddings. \mbox{``Sph-''} and ``Hyp-'' indicate Euclidean embeddings optimized using $D_{cos}$ (\ref{eq:dcos}) and hyperbolic embeddings optimized via $D_{hyp}$ (\ref{eq:hdist}). Models shown at the top employ ResNet-50~\cite{resnet} encoder, except IRT\textsubscript{R} is based on DeiT~\cite{deit}. In our method, we take $\lambda=3$ for CUB, and 8 for Cars and SOP. \hspace{\textwidth} \textsuperscript{$\mathsection$} models pretrained on the larger ImageNet-21k~\cite{imagenet}. The best results, within a non-significant difference of 0.1, are highlighted.
}  
  \label{tab:exp_128}
\end{table*}

\setlength{\tabcolsep}{0.35em}
\begin{table*}[htbp]
  \centering
  \small
  \scalebox{.8}{
  \begin{tabular}{l|c|cccc|cccc|cccc}
    \toprule
    \multirow{2}{*}{Method} &
    \multirow{2}{*}{Dim} &
    \multicolumn{4}{c|}{CUB-200-2011 (K)} &
    \multicolumn{4}{c|}{Cars-196 (K)} &
    \multicolumn{4}{c}{SOP (K)}\\
    & &
    1 & 2 & 4 & 8 &
    1 & 2 & 4 & 8 &
    1 & 10 & 100 & 1000\\
    \midrule
    A-BIER \cite{A_BIER} & 512 &
    57.5 & 68.7 & 78.3 & 86.2 &
    82.0 & 89.0 & 93.2 & 96.1 &
    74.2 & 86.9 & 94.0 & 97.8\\
    ABE \cite{ABE} & 512 &
    60.6 & 71.5 & 79.8 & 87.4 &
    85.2 & 90.5 & 94.0 & 96.1 &
    76.3 & 88.4 & 94.8 & 98.2\\
    SM \cite{SM} & 512 &
    56.0 & 68.3 & 78.2 & 86.3 &
    83.4 & 89.9 & 93.9 & 96.5 &
    75.3 & 87.5 & 93.7 & 97.4 \\
    XBM \cite{XBM} & 512 &
    65.8 & 75.9 & 84.0 & 89.9 &
    82.0 & 88.7 & 93.1 & 96.1  &
    79.5 & 90.8 & 96.1 & 98.7\\
    HTL \cite{HTL} & 512 &
    57.1 & 68.8 & 78.7 & 86.5 &
    81.4 & 88.0 & 92.7 & 95.7 &
    74.8 & 88.3 & 94.8 & 98.4\\
    MS \cite{MS} & 512 &
    65.7 & 77.0 & 86.3 & 91.2 &
    84.1 & 90.4 & 94.0 & 96.5 &
    78.2 & 90.5 & 96.0 & 98.7 \\
    SoftTriple \cite{softtriple} & 512 &
    65.4 & 76.4 & 84.5 & 90.4 &
    84.5 & 90.7 & 94.5 & 96.9 &
    78.6 & 86.6 & 91.8 & 95.4 \\
    HORDE \cite{HORDE} & 512 &
    66.8 & 77.4 & 85.1 & 91.0 &
    86.2 & 91.9 & 95.1 & 97.2 &
    80.1 & 91.3 & 96.2 & 98.7  \\
    Proxy-Anchor \cite{proxy_anchor} & 512 &
    68.4 & 79.2 & 86.8 & 91.6 &
    86.1 & 91.7 & 95.0 & 97.3 &
    79.1 & 90.8 & 96.2 & 98.7 \\
    NSoftmax \cite{NSoftmax} & 512 &
    61.3 & 73.9 & 83.5 & 90.0 &
    84.2 & 90.4 & 94.4 & 96.9 &
    78.2 & 90.6 & 96.2 & -  \\
    ProxyNCA++ \cite{PNCAPP} & 512 &
    69.0 & 79.8 & 87.3 & 92.7 &
    86.5 & 92.5 & 95.7 & 97.7 &
    80.7 & 92.0 & 96.7 & 98.9 \\
    IRT\textsubscript{R} \cite{IRT} & 384 &
    76.6 & 85.0 & 91.1 & 94.3 &
    - & - & - & - &
    84.2 & 93.7 & 97.3 & 99.1 \\
    \midrule
    ResNet-50 \cite{resnet} \textsuperscript{\dag} & 2048 &
    41.2 & 53.8 & 66.3 & 77.5 &
    41.4 & 53.6 & 66.1 & 76.6 &
    50.6 & 66.7 & 80.7 & 93.0 \\
    DeiT-S \cite{deit} \textsuperscript{\dag} & 384 &
    70.6 & 81.3 & 88.7 & 93.5 &
    52.8 & 65.1 & 76.2 & 85.3 &
    58.3 & 73.9 & 85.9 & 95.4 \\
    DINO \cite{dino} \textsuperscript{\dag} & 384 &
    70.8 & 81.1 & 88.8 & 93.5 &
    42.9 & 53.9 & 64.2 & 74.4 &
    63.4 & 78.1 & 88.3 & 96.0\\
    ViT-S \cite{vit_pretrain} \textsuperscript{\dag} \textsuperscript{$\mathsection$} & 384 &
    83.1 & 90.4 & 94.4 & 96.5 &
    47.8 & 60.2 & 72.2 & 82.6 &
    62.1 & 77.7 & 89.0 & 96.8 \\
    \midrule
    Sph-DINO & 384 &
    80.5 & 87.9 & 92.5 & 95.4 &
    87.0 & 92.7 & 95.9 & 97.6 & 
    84.8 & 94.2 & 97.6 & 99.2 
 \\
    Sph-ViT \textsuperscript{$\mathsection$} & 384 &
    {\bf 85.1} & 91.1 & 94.3 & 96.5 & 
    82.2 & 89.0 & 93.3 & 96.2 & 
    85.3 & 94.5 & 97.9 & {\bf 99.4} 
 \\
    Hyp-DINO & 384 &
    80.6 & 88.1 & 92.8 & 95.4 &
    87.5 & 93.1 & 96.2 & 97.9 &
    84.5 & 94.1 & 97.6 & 99.2 
 \\
    Hyp-ViT \textsuperscript{$\mathsection$} & 384 &
    85.0 & {\bf 91.3} & 94.5 & 96.5 & 
    84.3 & 90.8 & 94.8 & 97.3 & 
    85.2 & 94.6 & 98.0 & 99.4 
    \\
    Mix-DINO  & 384 &
    80.4 & 88.3 & 92.8 & 95.5 &
    {\bf 88.9} & {\bf 94.1} & {\bf 96.5} & {\bf 98.1} &
    85.1 & 94.5 & 97.8 & 99.3
    \\
    Mix-ViT \textsuperscript{$\mathsection$} & 384 &
    84.6 & {\bf 91.3} & {\bf 94.9} & {\bf 96.7} & 
    84.4 & 90.6 & 94.9 & 97.1 & 
    {\bf 86.0} & {\bf 95.0} & {\bf 98.2} & {\bf 99.5}
    \\
    \bottomrule
  \end{tabular}
  }
  \caption{
    Recall@K metric for three datasets. ``Dim'' indicates the dimensionality of embeddings. Two versions of our method, starting with ``Mix'', are evaluated for encoder embeddings and are listed in the bottom section. 
    \textsuperscript{\dag} pretrained encoders without training on the target dataset. \textsuperscript{$\mathsection$} means the model is pretrained on the larger ImageNet-21k \cite{imagenet}. The best results and the second best, within a difference of 0.1, are highlighted.
        }  
\vspace{-5pt}
  \label{tab:exp_big}
\end{table*}

\section{Experiments and Results}

We follow the training and evaluation protocol by \cite{ermolov2022hyperbolic}. We then contrast two versions of our approach with the current state-of-the-art across three benchmark datasets for image retrieval tasks. In the upcoming sections, we introduce the datasets and contrastive methods in Section \ref{exp:data}, provide implementation specifics in Section \ref{exp:detail}, and present the experimental results in Section \ref{exp:result}.

\subsection{Datasets}\label{exp:data}
We perform the evaluation of our method on three different benchmark datasets. 
{\bf CUB-200-2011} (CUB) \cite{cub200} comprises 11,788 images categorized into 200 different bird breeds. For our experiments, we utilize the first 100 classes containing 5,864 images as the training set, while the remaining 100 classes containing 5,924 images serve as the test set. The dataset poses a challenge due to the visual similarity between its instances; certain breeds can only be differentiated by subtle details. This aspect makes the dataset both challenging and informative for tasks such as image retrieval.
{\bf Cars-196} (Cars) \cite{cars196} consists of 16,185 images depicting 196 different car models. We allocate the first 98 classes (8,054 images) for training purposes and the remaining 98 classes (8,131 images) for testing.
{\bf Stanford Online Product} (SOP) \cite{sop} comprises 120,053 images depicting 22,634 products sourced from eBay.com. We adopt the standard train/test split, with 11,318 classes (59,551 images) designated for training and the remaining 11,316 classes (60,502 images) for testing.

\subsection{Implementation Details}\label{exp:detail}

We adopt the code implementation\footnote{\url{https://github.com/htdt/hyp_metric}} and hyper-parameter of \cite{ermolov2022hyperbolic}. For the pretrained backbone encoders, we use ViT-S \cite{vit_pretrain} with two different types of pretraining (ViT-S and DINO). The ViT architecture was introduced by \cite{vit}. The input image is divided into $16 \times 16$ patches, each of which is flattened. Subsequently, both the patch and its location information are linearly projected into an embedding. ViT-S \cite{vit_pretrain} is the smaller version of ViT with 6 heads in multiheaded self-attention (base version uses 12 heads). A detailed description is available in \cite{vit_pretrain}. We use the publicly available version of ViT-S pretrained on ImageNet-21k \cite{vit_pretrain}. The second encoder used in our experiments is DINO \cite{dino}, which is an architecture proposed for contrastive self-supervised representation learning. In this configuration, the model ViT-S is trained on the ImageNet-1k dataset \cite{imagenet1k} without labels. Detailed architecture design could be found in \cite{dino}. 

We replicate the experiment design and implementation details from previous work. Head biases are initialized to $0$, and weights with a (semi) orthogonal matrix \cite{ortho_init}. In fine-tuning, linear projection for patch embeddings remains frozen. The encoder yields a $384$ dimensional representation, projected by a head to a $128$ dimensional space. The model has two branches: one outputs a $128$ dimensional embedding in Euclidean space, while the other projects into hyperbolic space. Our proposed method for feature mixtures across geometries is then applied. Notably, encoder outputs are normalized prior to branching. Following \cite{ermolov2022hyperbolic}, we adopt naming conventions: the hyperbolic head baseline is ``Hyp-'', and the unit hypersphere is ``Sph-''.

Hyperbolic space consistently follows a trend when adjusting curvature parameter $c$ from small to large values. Optimal performance is achieved at $\tau=0.2$ then gradually declines. For smaller $c$ values, hyperbolic embeddings outperform Euclidean embeddings when $\tau > 0.2$. In our mixture model, we utilize $c=0.1$ and $\tau=0.2$ for hyperbolic embeddings, and $\tau=0.05$ for hypersphere embeddings. Lambda is tested from $1$ to $10$ to determine the best value. The clipping radius remains consistent with previous work at $r=2.3$.

We utilize the AdamW optimizer \cite{adamw} with a learning rate of $1 \times 10^{-5}$ for DINO and $3 \times 10^{-5}$ for ViT-S. The weight decay value is set to $0.01$, and the batch size is fixed at $900$. The number of optimizer steps varies depending on the dataset: $200$ for CUB, $600$ for Cars, and $25000$ for SOP. To ensure stability, we clip the gradient by norm $3$. Common data augmentations, including random crop resizing to $224 \times 224$ using bicubic interpolation and a random horizontal flip, are applied. We implement automatic mixed precision training using PyTorch \cite{paszke2019pytorch}. All experiments are conducted on a single NVIDIA A100 80G GPU.

For the baseline method proposed in \cite{ermolov2022hyperbolic}, we re-implemented the code and report the results run by ourselves.

\subsection{Results}\label{exp:result}

Tables \ref{tab:exp_128} and \ref{tab:exp_big} present the experimental recall@K metrics for the 128 dimensional head embedding and the results for the 384 dimensional encoder embedding. The evaluation results of the pretrained encoders without training on the target dataset in Tab. \ref{tab:exp_big} are sourced from \cite{ermolov2022hyperbolic}. The last 6 rows in the tables present the results obtained from our implementation.

In Fig. \ref{fig:bar}, it is shown that the performance of Euclidean embedding is governed by temperature parameter $\tau$. The smallee the temperature is, the better the performance. On the other hand, the performance of hyperbolic embedding is dominated by the choice of both $\tau$ and $c$. Different $c$ values will assign different relative distances between samples. Unlike the Euclidean embedding, most hyperbolic embedding shows better performance when $\tau$ is around 0.2. Different embeddings will give supplementary features to each other.

The observations from Fig.\ref{fig:bar}, Tab.\ref{tab:exp_128}, and Tab.~\ref{tab:exp_big} demonstrate that the Euclidean embedding model performs comparably, and sometimes even better, than the hyperbolic embedding method when the temperature parameter is small ($\tau=0.05$). This aligns with our analysis that Euclidean space provides sufficient embeddings as long as the temperature $\tau$ effectively promotes meaningful hard negative selection. Existing work lacks a comprehensive Euclidean vs. hyperbolic space comparison concerning temperature effects in the contrastive loss, leading to potentially inflated performance claims for hyperbolic embeddings.

For 128 dimensional embeddings, our Mix model achieves the best 1K recall value, outperforming Sph and Hyp baselines, even after their optimal tuning. In the CUB dataset, recall improvements for 1K, 2K, and 8K are 0.36\%, 0.22\%, and 0.21\%, respectively. In the Cars dataset, DINO consistently outperforms ViT, allowing our Mix method to also excel when using DINO as the backbone encoder. The relative improvements for 1K, 2K, 4K, and 8K are 2.40\%, 1.33\%, 0.85\%, and 0.52\%, respectively. Similarly, for SOP, there are relative improvements of 0.35\% and 0.11\% in 1K and 10K recall, respectively.

For 384 dimensional embeddings, our proposed Mix model shows promise. In the CUB dataset, 1K recall is slightly lower than the Sph-ViT method. Notably, on CUB, relative performance improvements for 4K and 8K are 0.42\% and 0.21\%, respectively. For the Cars dataset, relative improvements in 1K, 2K, 4K, and 8K recall are 1.60\%, 1.07\%, 0.31\%, and 0.20\%, respectively. In the SOP dataset, the relative improvements for 1K, 10K, 100K, and 1000K recall are 0.82\%, 0.42\%, 0.20\%, and 0.10\%, respectively. The experiments were conducted multiple times, with standard deviations around 0.2.

In our analysis in Section \ref{sec:method-mix}, we demonstrate that the value of $p(\mathbf x^-)$ varies between Euclidean and hyperbolic embeddings for the same instance. This discrepancy implies that the models will exhibit different preferences when selecting the corresponding triplet $(\mathbf x, \mathbf x^+, \mathbf x^-)$ to learn decision boundaries. We hypothesize that mixed geometry embeddings may not consistently outperform single geometry embeddings to a significant extent. In some cases, their performance could be comparable to that of single geometry embeddings. This is due to the fact that the selection of negatives in each geometry is determined by $p(\mathbf x^-)$. Instances with low values of $p(\mathbf x^-)$ in both geometries have minimal impact on the decision boundary. On the other hand, instances with high $p(\mathbf x^-)$ in one geometry but low in the other might be overlooked by one embedding, yet highlighted when the corresponding $p(\mathbf x^-)$ remains high after fusion. However, it's not guaranteed that the mixed model will always put emphasis on true hard negatives when learn a model. Our proposed method aims to validate our understanding of different geometry embeddings, rather than proposing an explicit negative selection method. This explains why the mixed model occasionally surpasses single-geometry embeddings but not consistently.

In our proposed method, we only tested simple addition as feature fusion. However, considering different $c$ values will result in new hyperbolic geometries, and further gains are possible if more embedding spaces are considered in the proposed Mix model.

\section{Conclusion}
Upon scrutinizing recent advancements in metric learning employing hyperbolic space, we uncover that the performance boost attributed to hyperbolic space primarily arises from parameter tuning of $\tau$ in the pairwise cross-entropy loss. 
Experimentally, we find Euclidean and hyperbolic embeddings exhibit complementary behavior with diverse $\tau$ and curvature values. This observation motivates us to delve deeper into hyperbolic metric learning. Our analysis reveals that the distinct performance of embeddings stems from varying effects of hard negative sampling in the two geometries, illustrated in Fig.~\ref{fig:negative}. To harness the benefits of these diverse geometries, we propose an ensemble learning approach that amalgamates features acquired in two spaces. By altering the hard negative data pool during the learning process, we encourage the model to acquire more informative features.

\bfsection{Limitation.}
In our current fusion experiment, we determine the optimal $\lambda$ value through a uniform search spanning 1 to 10. For future endeavors, an automated approach, such as incorporating $\lambda$ as a learnable parameter, could be explored. Additionally, our fusion model currently involves a simple convex combination of Euclidean and hyperbolic geometry. Considering the varying nature of hyperbolic geometry with changes in $c$, the fusion model's scope could potentially encompass more than two geometries. We believe this work offers insightful contributions to comprehending hyperbolic metric learning. Furthermore, the exploration and analysis of potential ensemble learning involving Euclidean and hyperbolic geometry in different configurations remain open avenues for future research.

\section*{Acknowledgements}

Yun Yue and Dr. Ziming Zhang were partially supported by NSF grant CCF-2006738. Fangzhou Lin was supported in part by the Support for Pioneering Research Initiated by the Next Generation (SPRING) from the Japan Science and Technology Agency.

{\small
\bibliographystyle{ieee_fullname}
\bibliography{egbib}
}











\setcounter{figure}{3}

\section*{A. Derivation  of Gradient in Eq. 8}

\begin{equation}\label{eq:grad}\small \tag{11}
\resizebox{0.88\hsize}{!}{%
$\begin{split}
    & \nabla\mathcal{L}_{NCE_i} \\
    & = \frac{1}{\left(  1 + \sum_{k \neq i} \text{exp} \left( \frac{1}{\tau}\left( s_{i, k} - s_{i, i}\right) \right)\right)} \nabla \sum_{k \neq i} \text{exp} \left( \frac{1}{\tau}\left( s_{i, k} - s_{i, i}\right) \right) \\
    & = \frac{1}{\tau} \frac{\sum_{k \neq i} \text{exp} \left( \frac{1}{\tau}\left( s_{i, k} - s_{i, i}\right) \right) \nabla \left( s_{i, k} - s_{i, i}\right)} {\left(  1 + \sum_{k \neq i} \text{exp} \left( \frac{1}{\tau}\left( s_{i, k} - s_{i, i}\right) \right)\right)} \\
    & = \frac{1}{\tau} \sum_{k \neq i} \frac{\text{exp} \left( \frac{1}{\tau} s_{i, k} \right )}{ \text{exp} \left( \frac{1}{\tau} s_{i, i} \right )  \left(  1 + \sum_{k \neq i} \text{exp} \left( \frac{1}{\tau}\left( s_{i, k} - s_{i, i}\right) \right)\right) } \nabla \left( s_{i, k} - s_{i, i}\right) \\
     & = \frac{1}{\tau} \sum_{k \neq i} \frac{\text{exp} \left(  s_{i, k} / \tau \right )}{\text{exp} \left(  s_{i, i} / \tau \right ) + \sum_{k \neq i} \text{exp} \left(  s_{i, k} / \tau \right )} \nabla \left( s_{i, k} - s_{i, i}\right)
\end{split}$
}
\end{equation}

\section*{B. Comparison with Different ViTs}

\begin{figure*}[b]
\small
\begin{center}
   \includegraphics[width=0.8\linewidth]{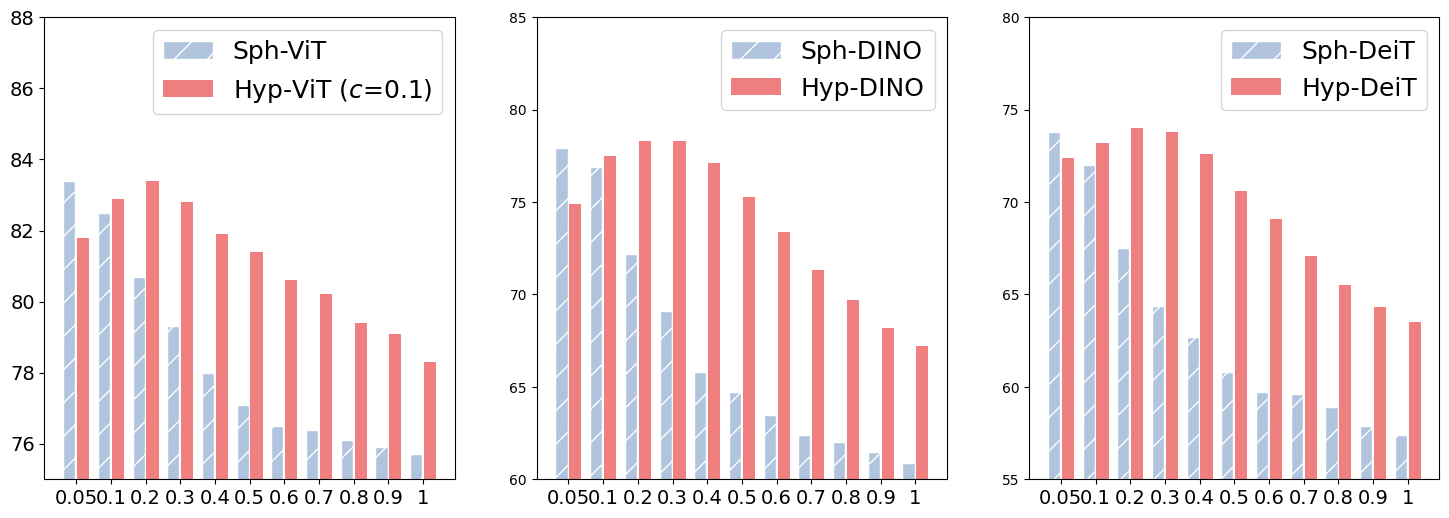}
\end{center}
   \caption{Recall of 1K metric comparison of models trained with different temperatures $\tau$ using CUB-200-2011 dataset. The x-axis indicates different $\tau$. \mbox{``Sph-''} are versions with hypersphere embeddings optimized using $D_{cos}$, ``Hyp-'' are versions with hyperbolic embeddings optimized using $D_{hyp}$. For ``Hyp-'' we fix the curvature parameter $c = 0.1$}
\label{fig:3models}
\end{figure*}

In Figure~\ref{fig:3models}, we demonstrate that temperature $\tau$ has a consistent impact on various ViT models when equipped with general contrastive loss and hyperbolic contrastive loss. Specifically, we evaluate ViT-s, DINO, and DeiT-s. Across different backbone transformer settings, hyperbolic embeddings consistently outperform Euclidean embeddings when $\tau>0.2$. For DINO hyperbolic embeddings show similar performance when $\tau=0.2$ and $\tau=0.3$. When $\tau$ increases, the performance of both Euclidean and hyperbolic embeddings drops. However, hyperbolic embeddings are always superior to the Euclidean case.  

\clearpage



\end{document}